\documentclass{article}

\usepackage[preprint]{neurips_2019}

\usepackage[utf8]{inputenc} 
\usepackage[T1]{fontenc}    
\usepackage{hyperref}       
\usepackage{url}            
\usepackage{booktabs}       
\usepackage{amsfonts}       
\usepackage{nicefrac}       
\usepackage{microtype}      
\usepackage{amsmath}
\usepackage{amssymb}

\title{Training Restricted Boltzmann Machines with Binary Synapses using the Bayesian Learning Rule}

\author{%
  Xiangming Meng\thanks{Most work performed when X. Meng was a postdoctoral researcher at RIKEN Center for Advanced Intelligence Project (AIP), Tokyo, Japan.} \\
  Institute for Physics of Intelligence\\
  The University of Tokyo\\
  Tokyo, Japan \\
  \texttt{meng@g.ecc.u-tokyo.ac.jp} \\
}

\begin{document}

\maketitle

\begin{abstract}
Restricted Boltzmann machines (RBMs) with low-precision synapses are much appealing with high energy efficiency. However, training RBMs with binary synapses is challenging due to the discrete nature of synapses. Recently \citet{huang2019data} proposed one efficient method to train RBMs with binary synapses by using a combination of gradient ascent and the message passing algorithm under the variational inference framework. However, additional heuristic
clipping operation is needed. In this technical note, inspired from  \citet{huang2019data} , we propose one alternative optimization method using the Bayesian learning rule, which is one natural gradient variational inference method. As opposed to \citet{huang2019data}, we update the natural parameters of the variational symmetric Bernoulli distribution rather than the expectation parameters. Since the natural parameters take values in the entire real domain, no additional clipping is needed.  Interestingly, the algorithm in \cite{huang2019data} could be viewed as one first-order approximation of the proposed algorithm, which justifies its efficacy with heuristic clipping. 
\end{abstract}

\section{Problem Formulation}
Restricted Boltzmann machines (RBMs) with low-precision discrete synapses are much appealing due to high energy efficiency. However, compared to full-precision RBMs, they are more difficult to train, which is essentially a discrete optimization problem. In a recent paper \cite{huang2019data}, the author addressed the problem of training RBMs with binary synaptic connections. The problem is formulated as follows. Consider RBMs where the random
visible variables $\mathbf{v}=\left\{ v_{1},...,v_{N}\right\} $ and
hidden variables $\mathbf{h}=\left\{ h_{1},...,h_{M}\right\} $ only
take binary values $\left\{ -1,+1\right\} $. Then the joint distribution
of this RBM model is given by the Gibbs distribution
\begin{equation}
p\left(\mathbf{v},\mathbf{h}\right)=\frac{1}{\mathtt{Z}}e^{-\beta E\left(\mathbf{v},\mathbf{h}\right)},\label{eq:Gibbs_distrib}
\end{equation}
where $\mathsf{Z}$ is the normalization constant, $\beta$ is the
temperature value, and $E\left(\mathbf{v},\mathbf{h}\right)$ is the
energy function defined as 
\begin{equation}
E\left(\mathbf{v},\mathbf{h}\right)=-\sum_{\mu=1}^{M}\sum_{i=1}^{N}w_{\mu i}h_{\mu}v_{i}-\sum_{i=1}^{N}b_{i}v_{i}-\sum_{\mu=1}^{M}c_{\mu}h_{\mu}.\label{eq:Energy_function}
\end{equation}

For simplicity and without loss of generality,  assume a simple case where the biases $b_{i}=0,i=1...N$
and $c_{\mu}=0,\mu=1...M$. The marginal distribution of $\mathbf{v}$
could be obtained by marginalizing out the hidden states $\mathbf{h}$
\begin{align}
p\left(\mathbf{v}\right) & =\frac{1}{Z\left(\mathbf{W}\right)}\prod_{\mu=1}^{M}\cosh\left(\beta X_{\mu}\right)\label{eq:prob_v}\\
X_{\mu} & \equiv\frac{1}{\sqrt{N}}\sum_{i=1}^{N}w_{\mu i}v_{i}=\frac{1}{\sqrt{N}}\mathbf{w}_{\mu}^{T}\mathbf{v}\label{eq:X_mu_def}
\end{align}
where $\mathbf{w}_{\mu}^{T}$ is the $\mu\textrm{-th}$ row of the
synaptic connection matrix $\mathbf{W}$, $X_{\mu}$ is the receptive
field of the $\mu\textrm{-th}$ hidden neuron, and $Z\left(\mathbf{W}\right)=\sum_{\mathbf{v}}\prod_{\mu=1}^{M}\cosh\left(\beta X_{\mu}\right)$
is the partition function depending on the synaptic connection matrix
$\mathbf{W}$. 

When we have $D$ input data samples $\mathbb{D}=\left\{ \mathbf{v}_{a}\right\} _{a=1}^{D}$
which are weakly-correlated, then the likelihood distribution of data
could be written as
\begin{equation}
p\left(\mathbb{D}\mid\mathbf{W}\right)=\prod_{a=1}^{D}\frac{1}{Z\left(\mathbf{W}\right)}\prod_{\mu=1}^{M}\cosh\left(\beta X_{\mu}^{a}\right),\label{eq:likelihood_dist}
\end{equation}
where $X_{\mu}^{a}$ is the receptive field of the $\mu\textrm{-th}$
hidden neuron for the $a\textrm{-th}$ data sample $\mathbf{v}_{a}.$ 
From the Bayesian perspective, suppose that the prior distribution of $\mathbf{W}$ is $p_{0}\left(\mathbf{W}\right)$,
according to Bayes' rule, the posterior distribution could be obtained
as
\begin{equation}
p\left(\mathbf{W}\mid\mathbb{D}\right)=\frac{p\left(\mathbb{D}\mid\mathbf{W}\right)p_{0}\left(\mathbf{W}\right)}{p\left(\mathbb{D}\right)},\label{eq:post_def}
\end{equation}
where $p\left(\mathbb{D}\right)=\sum_{\mathbf{W}}p\left(\mathbb{D}\mid\mathbf{W}\right)p_{0}\left(\mathbf{W}\right)$
is the partition function of the posterior and also known as the marginal
data likelihood. 

The goal of training RBMs with binary synapses is to learn the synaptic connection matrix
$\mathbf{W}$ from the observed data samples $\mathbb{D}=\left\{ \mathbf{v}_{a}\right\} _{a=1}^{D}$, subject to the discrete constraint that each element $w_{\mu i}$ in $\mathbf{W}$ also takes binary value, i.e., $w_{\mu i} \in \left\{ -1,+1\right\}$. If the posterior distribution $p\left(\mathbf{W}\mid\mathbb{D}\right)$
could be computed, then the learning problem 
is solved. However, exact computation of $p\left(\mathbf{W}\mid\mathbb{D}\right)$
is intractable.

For RBMs with full-precision synaptic connections, some classical training methods have been proposed such as the contrastive divergence (CD) algorithm \cite{hinton2002training}. However, in the case of RBMs with binary synaptic connections, it is essentially a challenging discrete optimization problem. As a result, the previous full-precision learning algorithms such as CD could not be used due to the discrete nature of the synapses.   

\section{Review of Huang's Method in \cite{huang2019data}}
Recently, \cite{huang2019data} addressed this challenging problem using a combination of
gradient ascent \footnote{It could be also equivalently understood as minimizing the negative ELBO using gradient descent (GD).} and the message passing algorithm under the variational inference (VI) framework. 
Specifically,
instead of computing the posterior directly, VI tries to find an approximate
distribution $q_{\boldsymbol{\lambda}}\left(\mathbf{W}\right)$ that
maximizes a lower bound of the log marginal likelihood $\log p\left(\mathbb{D}\right)$,
which is called the evidence lower bound (ELBO), i.e.,
\begin{equation}
\mathcal{L}\left(q_{\boldsymbol{\lambda}}\right)=\mathbb{E}_{q_{\boldsymbol{\lambda}}\left(\mathbf{W}\right)}\left[\log p\left(\mathbb{D}\mid\mathbf{W}\right)\right]-KL\left(q_{\boldsymbol{\lambda}}\left(\mathbf{W}\right)\parallel p_{0}\left(\mathbf{W}\right)\right),\label{eq:ELBO_def}
\end{equation}
where $KL\left(q\parallel p\right)=\mathbb{E}_{q}\left[\log\frac{q}{p}\right]$
is the Kullback-Leibler (KL) divergence and $p_{0}\left(\mathbf{W}\right)$ is the prior distribution which is assumed to be factorized as
\begin{align}
p_{0}\left(\mathbf{W}\right) & =\prod_{\mu=1}^{M}\prod_{i=1}^{N}\left[\frac{1+m_{\mu i}}{2}\delta\left(w_{\mu i}=1\right)+\frac{1-m_{\mu i}}{2}\delta\left(w_{\mu i}=-1\right)\right],\label{eq:prior_bernoulli_dist}
\end{align}
where $m_{\mu i}$ is the prior mean of $w_{\mu i}$ and also controls the probability $p(w_{\mu i}=+1)=(1+m_{\mu i})/2$. In practice, it is usually assumed that $m_{\mu i}=0$ when no informative prior information is available about the synapses. 
Alternatively, $\mathcal{L}\left(q_{\boldsymbol{\lambda}}\right)$ in (\ref{eq:ELBO_def})
could be rewritten as
\begin{equation}
\mathcal{L}\left(q_{\boldsymbol{\lambda}}\right)=\log p\left(\mathbb{D}\right)-KL\left(q_{\boldsymbol{\lambda}}\left(\mathbf{W}\right)\parallel p\left(\mathbf{W}\mid\mathbb{D}\right)\right),\label{eq:ELBO_2}
\end{equation}
so that $\mathcal{L}\left(q_{\boldsymbol{\lambda}}\right)\leq\log p\left(\mathbb{D}\right)$ and maximizing $\mathcal{L}\left(q_{\boldsymbol{\lambda}}\right)$ is equivalent to minimizing the KL divergence $KL\left(q_{\boldsymbol{\lambda}}\left(\mathbf{W}\right)\parallel p\left(\mathbf{W}\mid\mathbb{D}\right)\right)$.  Hence, the problem of posterior inference problem in (\ref{eq:post_def}) is  transformed to the optimization of $\mathcal{L}\left(q_{\boldsymbol{\lambda}}\right)$ with respect to (w.r.t.) the variational parameters $\boldsymbol{\lambda}$ of $q_{\boldsymbol{\lambda}}\left(\mathbf{W}\right)$, which is the core of VI. 

To model the binary synaptic connections weights $\mathbf{W}$, in \cite{huang2019data} the variational distribution $q_{\boldsymbol{\lambda}}\left(\mathbf{W}\right)$ is chosen to be a mean-filed  symmetric Bernoulli distribution 
\begin{align}
q_{\boldsymbol{\lambda}}\left(\mathbf{W}\right) & =\prod_{\mu=1}^{M}\prod_{i=1}^{N}\left[\frac{1+\eta_{\mu i}}{2}\delta\left(w_{\mu i}=1\right)+\frac{1-\eta_{\mu i}}{2}\delta\left(w_{\mu i}=-1\right)\right],\label{eq:post_bernoulli_dist}
\end{align}
where $\eta_{\mu i}\in\left[-1,1\right]$ is the posterior mean of $w_{\mu i}$ and it controls the probability
of the value of binary synaptic connection $w_{\mu i}\in\left\{ -1,+1\right\} $,
i.e., the probability of $w_{\mu i}=1$ is $\frac{1+\eta_{\mu i}}{2}$
while the probability of $w_{\mu i}=-1$ is $\frac{1-\eta_{\mu i}}{2}$.

Then, \cite{huang2019data} uses gradient ascent to  update the variational parameters $\eta_{\mu i}$, i.e., in the $t$-th iteration, each parameter $\eta_{\mu i}$ is updated as
\begin{align}
\eta_{\mu i}^{t+1} & =\eta_{\mu i}^{t}+\alpha\nabla_{\eta_{\mu i}}\mathcal{L}\left(q_{\boldsymbol{\lambda}^{t}}\right),\label{eq:GD_update_rule}
\end{align}
which seems easy to implement as long as the gradient term $\nabla_{\eta_{\mu i}}\mathcal{L}\left(q_{\boldsymbol{\lambda}^{t}}\right)$ is obtained. 
However, in contrast to the case of supervised learning, it is far from trivial to obtain the  gradient  $\nabla_{\eta_{\mu i}}\mathcal{L}\left(q_{\boldsymbol{\lambda}^{t}}\right)$. To be clear,
according to   (\ref{eq:ELBO_def}), the gradient  consists of two terms 
\begin{align}
\nabla_{\eta_{\mu i}}\mathcal{L}\left(q_{\boldsymbol{\lambda}^{t}}\right) = \nabla_{\eta_{\mu i}}\mathbb{E}_{q_{\boldsymbol{\lambda}}\left(\mathbf{W}\right)}\left[\log p\left(\mathbb{D}\mid\mathbf{W}\right)\right] - \nabla_{\eta_{\mu i}}KL\left(q_{\boldsymbol{\lambda}}\left(\mathbf{W}\right)\parallel p_{0}\left(\mathbf{W}\right)\right). \label{eq:GD_twoterms}
\end{align}
The gradient of the KL regularization term could be easily computed as 
\begin{align}
\nabla_{\eta_{\mu i}}KL\left(q_{\boldsymbol{\lambda}}\left(\mathbf{W}\right)\parallel p_{0}\left(\mathbf{W}\right)\right) = - \sum_{x=\pm 1} \frac{x}{2} \left(\log \frac{1+xm_{\mu i}}{1+x\eta_{\mu i}} - 1 \right). \label{eq:GD_KL}
\end{align}
However, the gradient of the expected log-likelihood term is intractable as it involves the computation of another log partition function $ \log Z\left(\mathbf{W}\right)$, i.e.,  
\begin{equation}
\nabla_{\eta_{\mu i}}\mathbb{E}_{q_{\boldsymbol{\lambda}^{t}}\left(\mathbf{W}\right)}\left[\log p\left(\mathbb{D}\mid\mathbf{W}\right)\right]=\nabla_{\eta_{\mu i}}\mathbb{E}_{q_{\boldsymbol{\lambda}^{t}}\left(\mathbf{W}\right)}\left[\sum_{a=1}^{D}\sum_{\mu=1}^{M}\log\cosh\left(\beta X_{\mu}^{a}\right)-D\log Z\left(\mathbf{W}\right)\right].\label{eq:grad_expected_log}
\end{equation}

To address this problem, \cite{huang2019data} leverages the message passing algorithm to obtain an approximation of the log partition function.
Specifically,  as seen in (\ref{eq:X_mu_def}), each $X_{\mu}^{a}$ is a sum of a large number of nearly independent
random variables and hence, by the central limit theorem, follows
a Gaussian distribution $\mathcal{N}\left(X_{\mu}^{a};G_{\mu}^{a},\Xi_{\mu}^{2}\right)$,
where the mean and variance are defined as
\begin{align}
G_{\mu}^{a} & =\frac{1}{\sqrt{N}}\sum_{i=1}^{N}\eta_{\mu i}v_{i}^{a},\label{eq:Gu_mean}\\
\Xi_{\mu}^{2} & =\frac{1}{N}\sum_{i=1}^{N}\left(1-\eta_{\mu i}^{2}\right),\label{eq:Kesi_variance}
\end{align}

As a result, similar to the local reparameterization trick \cite{kingma2015variational}, the expected
log-likelihood could be approximated using the Monte-Carlo estimation
\begin{align}
 & \mathbb{E}_{q_{\boldsymbol{\lambda}^{t}}\left(\mathbf{W}\right)}\left[\sum_{a=1}^{D}\sum_{\mu=1}^{M}\log\cosh\left(\beta X_{\mu}^{a}\right)-D\log Z\left(\mathbf{W}\right)\right]\nonumber \\
\approx & \frac{1}{S_{1}}\sum_{a,\mu,s}\log\cosh\left(\beta G_{\mu}^{a}+\beta\Xi_{\mu}z_{\mu}^{s}\right)-\frac{D}{S_{2}}\sum_{s}\log\sum_{\boldsymbol{\sigma}}\prod\cosh\left(\beta G_{\mu}+\beta\Xi_{\mu}z_{\mu}^{s}\right),\label{eq:expected_log_MC}
\end{align}
where $z_{\mu}^{s}$ are samples drawn from standard normal distribution,
and $S_{1}$ and $S_{2}$ are the number of samples used to estimate
different terms of the expected log-likelihood, respectively. However,
even with MC sampling, the computation of expected log-likelihood
is still difficult due to the existence $\log\sum_{\bold{v}}\prod\cosh\left(\beta G_{\mu}+\beta\Xi_{\mu}z_{\mu}^{s}\right)$.
Interestingly. as pointed out in \cite{huang2019data}, the term $\log\sum_{\bold{v}}\prod\cosh\left(\beta G_{\mu}+\beta\Xi_{\mu}z_{\mu}^{s}\right)$
corresponds to the log partition function of an equivalent RBM whose
synaptic connections are $\eta_{\mu i}/\sqrt{N}$ and biases of hidden
neurons are $\Xi_{\mu}z_{\mu}^{s}$. As a result, the $\log\sum_{\bold{v}}\prod\cosh\left(\beta G_{\mu}+\beta\Xi_{\mu}z_{\mu}^{s}\right)$
could be efficiently computed by resorting to the message passing
algorithm. To this end, denote by $m_{i\rightarrow\mu}$ the messages from visible
neuron to hidden neuron and $u_{\mu\rightarrow i}$ the message from
hidden neuron to the visible neuron, respectively, then the message
passing equation reads
\begin{align}
m_{i\rightarrow\mu} & =\tanh\left(\sum_{v\in\partial i\setminus\mu}u_{v\rightarrow i}\right),\label{eq:message_v2h}\\
u_{\mu\rightarrow i} & =\tanh^{-1}\left(\tanh\left(\beta\chi_{\mu\rightarrow i}+\beta H_{\mu}\right)\tanh\left(\beta\frac{\eta_{\mu i}}{\sqrt{N}}\right)\right),\label{eq:message_h2v}
\end{align}
where
\begin{align}
\chi_{\mu\rightarrow i} & \equiv\frac{1}{\sqrt{N}}\sum_{j\in\partial\mu\setminus i}\eta_{\mu j}m_{j\rightarrow\mu},\\
H_{\mu} & \equiv\Xi_{\mu}z_{\mu}^{s}.
\end{align}

After a few iterations, the log partition function $\log Z\left(\mathbf{W}\right)$ could be obtained approximately and thus the gradient of expected log-likelihood
in (\ref{eq:expected_log_MC}) w.r.t. $\boldsymbol{\mathbf{\eta}}$ could be approximated as \citep{huang2019data} 
\begin{align}
& \nabla_{\eta_{\mu i}}\mathbb{E}_{q_{\boldsymbol{\lambda}^{t}}\left(\mathbf{W}\right)}\left[\log p\left(\mathbb{D}\mid\mathbf{W}\right)\right] \nonumber \\ 
\approx & \frac{\beta}{S_{1}\sqrt{N}}\sum_{a,s}v_{i}^{a}\tanh\left(\beta G_{\mu}^{a}+\beta\Xi_{\mu}z_{\mu}^{s}\right) -\frac{\beta^{2}\eta_{\mu i}}{S_{1}N}\sum_{a,s}\left[1-\tanh^{2}\left(\beta G_{\mu}^{a}+\beta\Xi_{\mu}z_{\mu}^{s}\right)\right]\nonumber \\
& -\frac{D\beta}{S_{2}\sqrt{N}}\sum_{s}\left[C_{\mu i}-\frac{\eta_{\mu i}z_{\mu}^{s}}{\sqrt{N}\Xi_{\mu}}\hat{m}_{\mu}\right],
\label{eq:grad_first_expectedlog}
\end{align}
where
\begin{align}
m_{i} & =\tanh\left(\sum_{\mu\in\partial i}u_{\mu\rightarrow i}\right)\\
\hat{m}_{\mu} & =\int Dz\tanh\left(\beta\tilde{\chi}_{\mu}+\beta H_{\mu}+\beta\tilde{\Lambda}_{\mu}z\right)\\
C_{\mu i} & =\hat{m}_{\mu}m_{i}+\beta\frac{\eta_{\mu i}}{\sqrt{N}}\left(1-m_{i}^{2}\right)B_{\mu}\\
B_{\mu} & =1-\int Dz\tanh^{2}\left(\beta\tilde{\chi}_{\mu}+\beta H_{\mu}+\beta\tilde{\Lambda}_{\mu}z\right)
\end{align}
and $Dz\equiv e^{-z^{2}/2}/\sqrt{2\pi}dz$, $\tilde{\chi}_{\mu}\equiv\frac{1}{\sqrt{N}}\sum_{j\in\partial\mu}\eta_{\mu i}m_{i}$,
and $\tilde{\Lambda}_{\mu}\equiv\frac{1}{N}\sum_{i\in\partial\mu}\eta_{\mu i}^{2}\left(1-m_{i}^{2}\right)$. 

Finally, the update equation in \citet{huang2019data} for the variational parameters $\eta_{\mu i}$ is 
\begin{align}
\eta_{\mu i}^{t+1} = &\eta_{\mu i}^{t}+\alpha \sum_{x=\pm 1} \frac{x}{2} \left(\log \frac{1+xm_{\mu i}}{1+x\eta_{\mu i}} - 1 \right) + \alpha \frac{\beta}{S_{1}\sqrt{N}}\sum_{a,s}v_{i}^{a}\tanh\left(\beta G_{\mu}^{a}+\beta\Xi_{\mu}z_{\mu}^{s}\right) \nonumber \\
& -\alpha \frac{\beta^{2}\eta_{\mu i}}{S_{1}N}\sum_{a,s}\left[1-\tanh^{2}\left(\beta G_{\mu}^{a}+\beta\Xi_{\mu}z_{\mu}^{s}\right)\right]-\alpha\frac{D\beta}{S_{2}\sqrt{N}}\sum_{s}\left[C_{\mu i}-\frac{\eta_{\mu i}z_{\mu}^{s}}{\sqrt{N}\Xi_{\mu}}\hat{m}_{\mu}\right].\label{eq:GD_update_rule_final}
\end{align}
Since $\eta_{\mu i}\in\left[-1,1\right]$, the update in (\ref{eq:GD_update_rule_final}) could not guarantee such constraint. As a result, similar to \cite{baldassi2018role}, a heuristic clipping operation is introduced in \cite{huang2019data}, which forces the $\eta^t_{\mu i}=1$ when $\eta^t_{\mu i}>1$ and $\eta^t_{\mu i}=-1$ when $\eta^t_{\mu i}<-1$. This trick is heuristic and but works well empirically. One natural question is that: are any principled explanations for the heuristic clipping operation? Or are there any other algorithms without such clipping operation? 

\section{Training RBMs with Binary Synapses using the Bayesian Learning Rule}
In this section, we propose one alternative method to train RBMs with binary synaptic connections using the Bayesian Learning Rule \cite{khan2017conjugate}, which is obtained by optimizing the variational objective by using natural gradient descent \cite{amari1998natural,hoffman2013stochastic,khan2017conjugate}. As demonstrated in \cite{emti2020bayesprinciple}, the Bayesian learning rule can be used to derive and justify many existing learning-algorithms in fields such as optimization, Bayesian statistics, machine learning and deep learning. Note that recently the Bayesian learning rule has been applied in \cite{meng2020training} to train binary neural networks for supervised learning. Therefore, this note could be viewed as an extension of \cite{meng2020training} to the case of unsupervised learning \footnote{However, despite using the same Bayesian learning rule, the resultant algorithm for unsupervised learning in this note is quite different from that in \cite{meng2020training} for supervised learning. }. 

Specifically, to optimize the variational objective in (\ref{eq:ELBO_def}), the Bayesian learning rule \cite{emti2020bayesprinciple} considers a class of minimal exponential family distribution
\begin{align}
q_{\boldsymbol{\lambda}}\left(\mathbf{W}\right) :=  h\left(\boldsymbol{\lambda}\right)\exp{\left[\boldsymbol{\lambda}^T\phi(\mathbf{W})-A(\boldsymbol{\lambda})\right]}
\label{eq:exponential_family}
\end{align}
where $\boldsymbol{\lambda}$ is the natural parameter, $\phi(\mathbf{W})$ is the vector of sufficient statistics, $A(\boldsymbol{\lambda})$ is the log-partition function, and $h\left(\mathbf{W}\right)$ is the base measure.
When the prior distribution $p_0(\mathbf{W})$ follows the same distribution as  $q_{\boldsymbol{\lambda}}\left(\mathbf{W}\right)$ in (\ref{eq:exponential_family}), and the base measure $h(\mathbf{W})=1$, the Bayesian learning uses the following update of the natural parameter \cite{emti2020bayesprinciple}
\begin{equation}
\boldsymbol{\lambda} \leftarrow (1-\alpha)\boldsymbol{\lambda} +  \alpha\left\{\nabla_{\boldsymbol{\eta}}\mathbb{E}_{q_{\boldsymbol{\lambda}}\left(\mathbf{W}\right)}\left[\log p\left(\mathbb{D}\mid\mathbf{W}\right)\right] + \boldsymbol{\lambda}_0\right\}, \label{eq:BayesLearningRule}
\end{equation}
where $\alpha$ is the learning rate,  $\boldsymbol{\eta}$ is the expectation parameter of $q_{\boldsymbol{\lambda}}\left(\mathbf{W}\right)$, and $\boldsymbol{\lambda}_0$ is the natural parameter of the prior distribution $p_0(\mathbf{W})$. The main idea is to update the natural parameters using the natural gradient. Below we briefly show how to obtain the Bayesian learning rule; for more details, please refer to \cite{emti2020bayesprinciple,khan2017conjugate}.

To apply the Bayesian learning rule, the posterior approximation $q_{\boldsymbol{\lambda}}\left(\mathbf{W}\right)$ is also chosen to be the fully factorized symmetric Bernoulli distribution in (\ref{eq:post_bernoulli_dist}), which is in fact belonging to the minimal exponential family distribution. In particular, $q_{\boldsymbol{\lambda}}\left(\mathbf{W}\right)$ in (\ref{eq:post_bernoulli_dist}) could be reformulated as follows 
\begin{align}
q_{\boldsymbol{\lambda}}\left(\mathbf{W}\right) & =\prod_{\mu=1}^{M}\prod_{i=1}^{N}\left(\frac{1+\eta_{\mu i}}{2}\right)^{\frac{1+w_{\mu i}}{2}}\left(\frac{1-\eta_{\mu i}}{2}\right)^{\frac{1-w_{\mu i}}{2}}\nonumber \\
 & =\prod_{\mu=1}^{M}\prod_{i=1}^{N}\exp\left\{ \frac{w_{\mu i}}{2}\log\left(\frac{1+\eta_{\mu i}}{1-\eta_{\mu i}}\right)+\frac{1}{2}\log\left(\frac{1-\eta_{\mu i}^{2}}{4}\right)\right\} \\
 & =\prod_{\mu=1}^{M}\prod_{i=1}^{N}\exp\left\{ \lambda_{\mu i}\phi\left(w_{\mu i}\right)-A\left(\lambda_{\mu i}\right)\right\} \\
 & \equiv\prod_{\mu=1}^{M}\prod_{i=1}^{N}q_{\lambda_{\mu i}}\left(w_{\mu i}\right),\label{eq:expo_distr}
\end{align}
where the natural parameter $\lambda_{\mu i}$, sufficient statistics
$\phi\left(w_{\mu i}\right)$ , log partition function $A\left(\lambda_{\mu i}\right)$
, and the associated expectation parameter $\eta_{\mu i}=\mathbb{E}_{q_{\lambda_{\mu i}}\left(w_{\mu i}\right)}\left[\phi\left(w_{\mu i}\right)\right]$
are as follows
\begin{align}
\lambda_{\mu i} & \equiv\frac{1}{2}\log\left(\frac{1+\eta_{\mu i}}{1-\eta_{\mu i}}\right)\label{eq:nat_param}\\
\phi\left(w_{\mu i}\right) & \equiv w_{\mu i}\label{eq:suff_stat}\\
A\left(\lambda_{\mu i}\right) & \equiv-\frac{1}{2}\log\left(\frac{1-\eta_{\mu i}^{2}}{4}\right)\label{eq:log_part}\\
\eta_{\mu i} & \equiv \tanh\left(\lambda_{\mu i}\right). \label{eq:expect_param}
\end{align}

As a result, instead of optimizing the expectation parameters $\eta_{\mu i}$ using gradient ascent in (\ref{eq:GD_update_rule}) as \cite{huang2019data}, we could update the natural parameters $\lambda_{\mu i}$ using the Bayesian learning rule in (\ref{eq:BayesLearningRule}). Interestingly, as shown in (\ref{eq:BayesLearningRule}), although the natural parameters $\lambda_{\mu i}$ are updated, the gradient is computed w.r.t. the expectation parameters $\eta_{\mu i}=\tanh\left(\lambda_{\mu i}\right)$, which is already obtained in (\ref{eq:grad_first_expectedlog}). When the prior $p_{0}\left(\mathbf{W}\right)$ is set to be the form in (\ref{eq:prior_bernoulli_dist}), each element of the natural parameters  $\boldsymbol{\lambda}_0$ could be written as
\begin{align}
\lambda^0_{\mu i} \equiv\frac{1}{2}\log\left(\frac{1+m_{\mu i}}{1-m_{\mu i}}\right).   
\end{align}
Therefore, substituting (\ref{eq:grad_first_expectedlog}) into (\ref{eq:BayesLearningRule}), the natural parameters $\lambda_{\mu i}$ 
could be updated as
\begin{align}
\lambda_{\mu i}^{t+1}  = & \lambda_{\mu i}^{t} + \alpha \left(\lambda^0_{\mu i}-\lambda_{\mu i}^{t}\right)+\alpha\frac{\beta}{S_{1}\sqrt{N}}\sum_{a,s}v_{i}^{a}\tanh\left(\beta G_{\mu}^{a}+\beta\Xi_{\mu}z_{\mu}^{s}\right)\nonumber \\
 & -\alpha\frac{\beta^{2}\eta_{\mu i}^{t}}{S_{1}N}\sum_{a,s}\left[1-\tanh^{2}\left(\beta G_{\mu}^{a}+\beta\Xi_{\mu}z_{\mu}^{s}\right)\right] -\alpha\frac{D\beta}{S_{2}\sqrt{N}}\sum_{s}[C_{\mu i}-\frac{\eta_{\mu i}^{t}z_{\mu}^{s}}{\sqrt{N}\Xi_{\mu}}\hat{m}_{\mu}]. \label{eq:natrual_update_final}
\end{align}

It is easy to verify that 
\begin{align}
\lambda^0_{\mu i}-\lambda_{\mu i}^{t} = \sum_{x=\pm 1} \frac{x}{2} \left(\log \frac{1+xm_{\mu i}}{1+x\eta_{\mu i}} - 1 \right). \label{eq:gradient_KL}
\end{align}
Note that there is no need in (\ref{eq:natrual_update_final}) to explicitly compute the right hand side term of (\ref{eq:gradient_KL}), which is different from (\ref{eq:GD_update_rule_final}). The resultant algorithm to train RBMs with binary synaptic connections with (\ref{eq:natrual_update_final}) is termed as Bayesian Binary RBMs (BayesBRBM). Note that in BayesBRBM, the update formula (\ref{eq:natrual_update_final}) is  similar to (\ref{eq:GD_update_rule_final}) used in \cite{huang2019data}. However, there are two fundamental differences. First, BayesBRBM updates the natural parameters $\lambda_{\mu i}$  of the symmetric Bernoulli distribution while \cite{huang2019data} updates the expectation parameters $\eta_{\mu i}$. One direct advantage is that since $\lambda_{\mu i} \in (-\infty,+\infty)$, no additional clipping operation is needed as \cite{huang2019data}. Second, although the update equations (\ref{eq:natrual_update_final}) and (\ref{eq:GD_update_rule_final}) appear the same, they actually correspond to two fundamentally different optimization methods: the former uses natural gradient ascent while the latter uses gradient ascent. 

Interestingly, the algorithm
in \cite{huang2019data} could be viewed as one kind of first-order approximation of BayesBRBM. Specifically, using first-order Taylor expansion, the expectation parameters $\eta_{\mu i}$ could be approximated as
\begin{align}
\eta_{\mu i}=\tanh\left(\lambda_{\mu i}\right) \approx \lambda_{\mu i}.
\label{eq:first-oder-approx}
\end{align}
Using the first-order approximation (\ref{eq:first-oder-approx}), the update equation in (\ref{eq:natrual_update_final}) is approximated as 
\begin{align}
\lambda_{\mu i}^{t+1}  = & \lambda_{\mu i}^{t} + \alpha \sum_{x=\pm 1} \frac{x}{2} \left(\log \frac{1+xm_{\mu i}}{1+x\lambda_{\mu i}} - 1 \right) +\alpha\frac{\beta}{S_{1}\sqrt{N}}\sum_{a,s}v_{i}^{a}\tanh\left(\beta G_{\mu}^{a}+\beta\Xi_{\mu}z_{\mu}^{s}\right)\nonumber \\
 & -\alpha\frac{\beta^{2}\lambda_{\mu i}^{t}}{S_{1}N}\sum_{a,s}\left[1-\tanh^{2}\left(\beta G_{\mu}^{a}+\beta\Xi_{\mu}z_{\mu}^{s}\right)\right] -\alpha\frac{D\beta}{S_{2}\sqrt{N}}\sum_{s}[C_{\mu i}-\frac{\lambda_{\mu i}^{t}z_{\mu}^{s}}{\sqrt{N}\Xi_{\mu}}\hat{m}_{\mu}], \label{eq:natrual_update_firstorder}
\end{align}
where the relation in (\ref{eq:gradient_KL}) is explicitly substituted for ease of comparison. It could be seen that the update formula in (\ref{eq:natrual_update_firstorder}) has exactly the same form as (\ref{eq:GD_update_rule_final}) except the exchange of variables between $\lambda_{\mu i}$ and $\eta_{\mu i}$. Since $\eta_{\mu i} \in [-1,+1]$, using first-order approximation (\ref{eq:first-oder-approx}), the values $\lambda_{\mu i}$ should also be constrained into the range $[-1,+1]$ by using clipping, which is exactly the algorithm in \cite{huang2019data}. As a result, the proposed algorithm provides a different perspective on \cite{huang2019data} which justifies its efficacy with heuristic clipping. 

\section{Summary}
In this technical note, building on the work in \citet{huang2019data},  we propose one optimization method called BayesBRBM (Bayesian Binary RBM) to train RBM with binary Synapses using the Bayesian learning rule. As opposed to \citet{huang2019data}, no additional clipping operation is needed for BayesBRBM. Interestingly, the method in \citet{huang2019data} could be viewed as a first-order approximation of BayesBRBM, which provides an alternative perspective and justifies its efficacy with heuristic clipping. One possible future work is to extend it to deep RBMs with binary synapses and make some detailed comparison of the two algorithms.

\subsubsection*{Acknowledgments}
X. Meng would like to thank Haiping Huang (Sun Yat-sen University) for helpful discussions, and Mohammad Emtiyaz Khan (RIKEN AIP) for explanations on the Bayesian learning rule.  

\bibliography{main}
\bibliographystyle{icml2020}

\section*{Appendix}
In this appendix, we briefly introduce the Bayesian learning rule. please refer to \cite{emti2020bayesprinciple,khan2017conjugate} for more details.
According to the definition of natural gradient ascent, the update
equation follows
\begin{equation}
\boldsymbol{\lambda}^{t+1}=\boldsymbol{\lambda}^{t}+\alpha\mathbf{F}\left(\boldsymbol{\lambda}^{t}\right)^{-1}\nabla_{\boldsymbol{\lambda}}\mathcal{L}\left(q_{\boldsymbol{\lambda}^{t}}\right)=\boldsymbol{\lambda}^{t}+\alpha\tilde{\nabla}_{\boldsymbol{\lambda}}\mathcal{L}\left(q_{\boldsymbol{\lambda}^{t}}\right),\label{eq:NGD_update_rule}
\end{equation}
where $\tilde{\nabla}_{\boldsymbol{\lambda}}\mathcal{L}\left(q_{\boldsymbol{\lambda}^{t}}\right)=\mathbf{F}\left(\boldsymbol{\lambda}^{t}\right)^{-1}\nabla_{\boldsymbol{\lambda}}\mathcal{L}\left(q_{\boldsymbol{\lambda}^{t}}\right)$
denotes the natural gradient of $\mathcal{L}\left(q_{\boldsymbol{\lambda}^{t}}\right)$
with respect to (w.r.t) $\boldsymbol{\lambda}$ at $\boldsymbol{\lambda}=\boldsymbol{\lambda}_{t}$,
where $\nabla_{\boldsymbol{\lambda}}\mathcal{L}\left(q_{t}\right)$
is the gradient of $\mathcal{L}\left(q\right)$ w.r.t $\boldsymbol{\lambda}$
at $\boldsymbol{\lambda}=\boldsymbol{\lambda}_{t}$ and $\mathbf{F}\left(\boldsymbol{\lambda}_{t}\right)$
is the Fisher information matrix (FIM) 
\begin{equation}
\mathbf{F}\left(\boldsymbol{\lambda}\right)\equiv\mathbb{E}_{q\left(\mathbf{w}\right)}\left[\nabla_{\boldsymbol{\lambda}}\log q_{\boldsymbol{\lambda}}\left(\mathbf{W}\right)\nabla_{\boldsymbol{\lambda}}\log q_{\boldsymbol{\lambda}}\left(\mathbf{W}\right)^{T}\right].
\end{equation}

As a result, to update natural parameters using the natural gradient we need to compute the inverse FIM, which is intractable in general. Fortunately,
for minimal exponential family distribution $q\left(\mathbf{W}\right)$ in
(\ref{eq:exponential_family}), there exists a concise result since
$\mathbf{F}\left(\boldsymbol{\lambda}^{t}\right)^{-1}\nabla_{\boldsymbol{\lambda}}\mathcal{L}\left(q_{\boldsymbol{\lambda}^{t}}\right)=\nabla_{\boldsymbol{\eta}}\mathcal{L}\left(q_{\boldsymbol{\lambda}^{t}}\right)$
where $\boldsymbol{\eta}$
is the expectation parameter of exponential family distribution $q_{\boldsymbol{\lambda}}\left(\mathbf{W}\right)$.
As a result, $\tilde{\nabla}_{\boldsymbol{\lambda}}\mathcal{L}\left(q_{\boldsymbol{\lambda}^{t}}\right)=\nabla_{\boldsymbol{\eta}}\mathcal{L}\left(q_{\boldsymbol{\lambda}^{t}}\right)$
so that the natural gradient update in (\ref{eq:NGD_update_rule}) could be equivalently
written as
\begin{align}
\boldsymbol{\lambda}^{t+1} & =\boldsymbol{\lambda}^{t}+\alpha\nabla_{\boldsymbol{\mu}}\mathcal{L}\left(q_{\boldsymbol{\lambda}^{t}}\right),\label{eq:NGD_update_rule-1}
\end{align}
where, from the definition of $\mathcal{L}\left(q_{\boldsymbol{\lambda}}\right)$
in (\ref{eq:ELBO_def}), there is 
\begin{align}
 \nabla_{\boldsymbol{\eta}}\mathcal{L}\left(q_{\boldsymbol{\lambda}^{t}}\right) =\nabla_{\boldsymbol{\eta}}\mathbb{E}_{q_{\boldsymbol{\lambda}^{t}}\left(\mathbf{W}\right)}\left[\log p\left(\mathbb{D}\mid\mathbf{W}\right)\right]-\left(\boldsymbol{\lambda}^{t}-\boldsymbol{\lambda}_{0}\right).\label{eq:gradient_mean}
\end{align}
Substituting (\ref{eq:gradient_mean}) into (\ref{eq:NGD_update_rule-1}) leads to the Bayesian learning rule in (\ref{eq:BayesLearningRule}).

\end{document}